\documentclass[5p]{elsarticle}
\usepackage{hyperref}

\journal{Journal of \LaTeX\ Templates}

\usepackage{algorithm}
\usepackage{algorithmic}
\usepackage{amsmath}
\usepackage{amsfonts}
\usepackage{epstopdf}

\usepackage{color}
\usepackage{hyperref}
\usepackage{graphicx}
\usepackage{subfigure}

\begin{document}	
	\begin{frontmatter}
		
		\title{Visual Object Tracking by Segmentation with Graph Convolutional Network}	
	
	\author[mymainaddress]{Bo Jiang}
	\cortext[mycorrespondingauthor]{Corresponding author. }
	\author[mymainaddress]{Panpan Zhang}	
	\author[mymainaddress]{Lili Huang\corref{mycorrespondingauthor}}	
	
	\address[mymainaddress]{Anhui Provincial Key Lab of Multimodal Cognitive Computation, School of Computer Science and Technology, Anhui University, Hefei, China}

\begin{abstract}			

Segmentation-based tracking has been actively studied in computer vision and multimedia. Superpixel based object segmentation and tracking methods are usually developed for this task. However, they independently perform feature representation and learning of superpixels which may lead to sub-optimal results. In this paper, we propose to utilize graph convolutional network (GCN) model for superpixel based object tracking. The proposed model provides a general end-to-end framework which integrates
i) label linear prediction, and ii) structure-aware feature information of each superpixel together to obtain object
segmentation and further improves the performance of tracking.
The main benefits of the proposed GCN method
have two main aspects. First, it provides an effective end-to-end way to exploit both spatial and temporal consistency constraint for target object segmentation. Second, it utilizes a mixed graph convolution  module to learn a context-aware and discriminative feature for superpixel representation and labeling. An effective algorithm has been developed to optimize the proposed model. Extensive experiments on five datasets demonstrate that our method obtains better performance against existing alternative methods.

%
		\end{abstract}
		
		
	\end{frontmatter}
	

\section{Introduction}

\begin{figure}[htb]
	\label{img::comparision}
	\subfigure[]{
		\begin{minipage}[t]{0.5\textwidth}
			\centering
			\includegraphics[width=\textwidth]{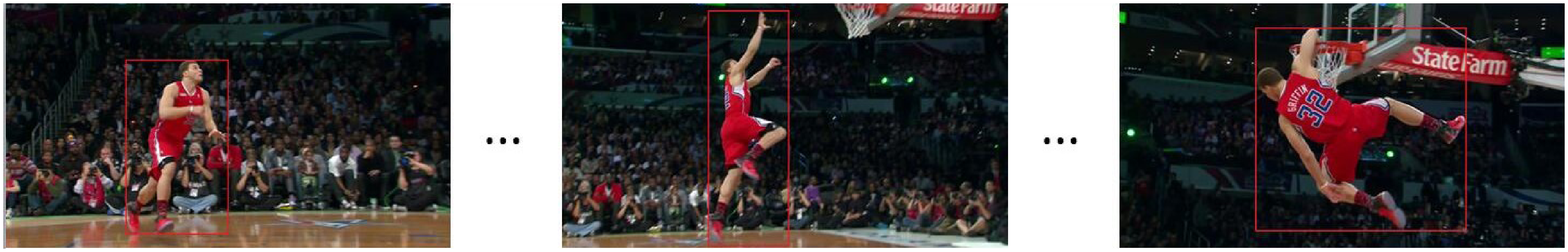}
		\end{minipage}
	}
\subfigure[]{
	\begin{minipage}[t]{0.5\textwidth}
		\centering
		\includegraphics[width=\textwidth]{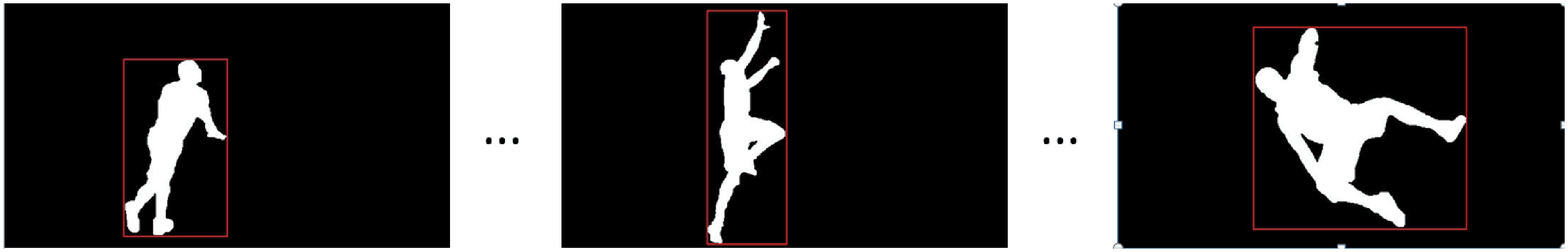}
   \end{minipage}
   }
\caption{Comparison of tracking-by-segmentation and tracking-by-detection. In a video sequence, as shown in (a), the detection-based tracking generally uses a bounding box to locate the target, while (b) illustrates that segmentation-based tracking segments the target and then locates the target according to the segmentation mask.}	
\end{figure}

Visual object tracking has attracted public attention due to
its extensive applications in the field of computer vision,
such as virtual reality, medical image classification and activity analysis.
%
 In general, most of existing methods can be categorized into two branches, i.e., tracking-by-detection and tracking-by-segmentation.
 Tracking-by-detection methods~\cite{MEEM2014,Babenko2011Robust,Grabner2008Semi,6619157,Danelljan2016ECO,Hong2015MUlti,Danelljan2014Accurate,Jiang2019Robust,HuangXLY2019,Feng2020Tracking}  aim to estimate the locations of the target object in the video sequences by using a rectangle bounding box around the target object,
as shown in Figure~\ref{img::comparision}(a).
In contrast, tracking-by-segmentation methods~\cite{Yeo2017Superpixel,Tracking-by-Segmentation2018} pixel-wisely segment the object from background and then track the target according to the segmentation results, as shown in Figure~\ref{img::comparision}(b).

Early tracking-by-segmentation works  usually adopt pixel-wise segmentation strategies~\cite{5539810,Godec2011Hough,Duffner2014PixelTrack}.
For example, Chad et al.~\cite{5539810} propose a probability framework based on the joint Gaussian distribution  over all the pixels to jointly solve the segmentation and tracking. Godec et al.~\cite{Godec2011Hough} use the generalized Hough transform and online learning technology to roughly separate the object and background pixel by pixel. Duffner et al.~\cite{Duffner2014PixelTrack} integrate generalized Hough transform with probabilistic segmentation approach together for pixel-based object tracking.
Although these segmentation methods can distinguish the foreground and background, their abilities to handle the heavily occlusion and clutter are limited because they fail to consider the internal structural information of the target object.
Besides, pixel-level processing usually has high computational complexity.
To alleviate these issues, some works utilize super-pixel methods to model non-rigid and deformable object tracking and reduce dimensions. Yang et al.~\cite{yang2014robust} propose a model for appearance discrimination, which extract middle-level features from hyper-pixels to distinguish target objects and background. Wang et al.~\cite{7875137} propose a constraint graph labeling algorithm, which take into account the internal structure information of the graph. The nodes of the graph model represent hyper-pixels, and the space, time and appearance fitness constraints of the bottom layer are encoded by the edge. Yeo et al.~\cite{Yeo2017Superpixel} propose to use Absorbing Markov Chain (AMC) model for super-pixel segmentation.
However, one main limitation of the above approaches is that they  usually individually conducts super-pixel feature representation and learning (or labeling) which may lead to sub-optimal results.
%

To overcome above issues, in this paper, we propose to utilize Graph Convolutional Network (GCN) model for super-pixel based object segmentation and tracking.
The proposed model provides a general framework that integrates the super-pixel feature representation and labeling together in an end-to-end framework.
The main advantage of the proposed GCN method is that it provides an effective context-aware representation method for each hyper-pixel by using the spatial-temporal structural information of different hyper-pixels.
Also, we provide an efficient way to implement the proposed model.
Overall, the main contributions of this paper are summarized as follows.

%
%

\begin{itemize}
	\item We propose to employ a graph convolutional network model to obtain context-aware feature representations for super-pixels which improves tracking performance.
\item We present to design an efficient and effective optimization algorithm to solve the proposed   model.
	\item Experiments on several widely used benchmark data-sets demonstrate the superiority of our model over current tracking approaches.
\end{itemize}

\section{Related Work}
\subsection{Graph convolutional network}

Graph convolutional networks(GCNs) have been commonly studied in computer vision field.
As a branch of GNNs, GCNs can extend the convolutional architecture to arbitrary graph-structural data including both regular and irregular structures~\cite{DuvenaudConvolutional,KipfSemi,NiepertLearning}. For example, M-GLCN~\cite{M-GLCN2019} is proposed for image co-saliency estimation task.~\cite{Multi-Label2019} demonstrates the advantages of GCNs on multi-label classification task. ~\cite{Qiu2018DeepInf} applies GCN to social influence prediction. PH-GCN~\cite{PH-GCN2019} is designed for person Re-ID problem. 
Park et al, ~\cite{ParkSymmetric} propose GALA for unsupervised graph representation and learning. 
Jiang et al, ~\cite{GLMNet2019} propose GLMNet for feature matching problem by incorporating both smoothing and resharpening together~\cite{ParkSymmetric}. 
Different from these works~\cite{GLMNet2019,ParkSymmetric}, we adapt both smoothing and resharpening GCN for visual object segmentation and tracking problem. Also, we develop a simple and effective graph network, followed by an efficient optimization algorithm, for our tracking task based on SGCN~\cite{WuSimplifying}.
Recently, GCN is also applied to  tracking-by-detection problems. Tu et al.~\cite{tu2019visual} propose to use GCNs to convert the heterogeneous features extracted from CNN into structured information. Gao et al.~\cite{GCT} integrate two different GCNs into existing Siamese methods~\cite{2016Fully} to model the appearance of an object fully considering the space-time information of the object context. However, the detection-based tracking methods cannot display the segmentation mask of the tracking target. In contrast, in this paper, we mainly focus on integrating GCN into the segmentation-based tracking framework.


\subsection{Tracking-by-segmentation}
Here, we briefly review some related works on segmentation based tracking problem~\cite{Godec2011Hough,OGBDT2015,Yeo2017Superpixel,Tracking-by-Segmentation2018}. In work~\cite{Godec2011Hough}, it extends Hough Forests to online mode and couples the rough GrabCut segmentation with voting-based detection and back-projection. Son et al.~\cite{OGBDT2015} propose an online tracking algorithm based on online gradient enhancement decision tree. Yeo et al.~\cite{Yeo2017Superpixel} present a tracking-by-segmentation framework using absorbing markov model to better distinguish the foreground and background. Lee et al.~\cite{Tracking-by-Segmentation2018} design an extended version of~\cite{Yeo2017Superpixel} by replacing the simple feature representation in ~\cite{Yeo2017Superpixel} with multi-level convolutional features. Different from  previous works, this paper  propose a contextual feature representations with stronger discrimination for super-pixels by using graph convolution representation model. In addition, we conduct feature learning and labeling together in an end-to-end network framework, which helps to accurately segment the target of each frame.


\section{The proposed Model}

Given a candidate region of the target object,
we first partition it into non-overlapping superpixels via Simple Linear Iterative Clustering (SLIC) algorithm~\cite{Achanta2012SLIC} and then
extract the feature descriptor for each superpixel.
Then, we aim to assign a weight value to represent its possibility of belonging to the target. To this end, we formulate this task as graph representation and labeling and propose
to employ a graph convolutional network model (GCN) for this task. The whole flow of the proposed network is shown in Figure~\ref{img::framework}.
%
\begin{figure*}[htb]
	\label{img::framework}
	\centering
	\includegraphics[scale=0.45]{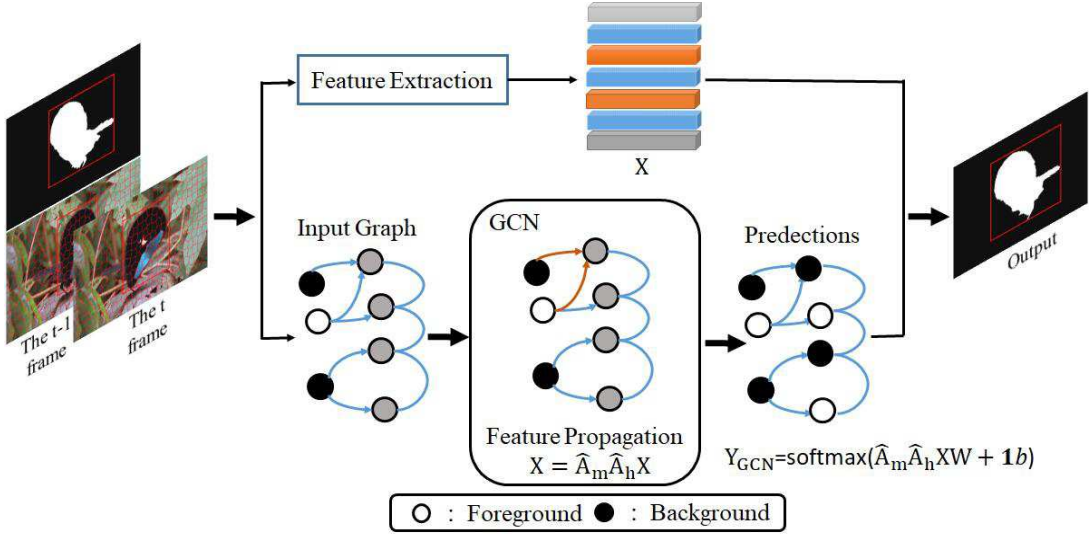}
	\caption{Architecture of the proposed GCN network for tracking-by-segmentation.}
	\label{img::framework}
\end{figure*}

\subsection{Graph construction}

Let $X=(X^{t-1}, X^{t})$ be the
collection of feature descriptors of super-pixels of two consecutive  frames t-1, t where $X^{t-1}\in \mathbb{R}^{d\times n_{t-1}}$ and $X^{t}\in \mathbb{R}^{d\times n_{t}}$ and let $ n_{t} $ denote the number of super-pixels in frame t.
%
We first construct a spatial-temporal graph as $G = (V,E)$,  where $V = \{V^{t-1},V^{t}\}$ denotes the superpixels node set of two consecutive frames and edge set $E$ represents the relationships among different superpixels. There are two kinds of edges in $G = (V,E)$, which represent the similarities among nodes within each frame and between different continuous frames.
%
First, the weight of edge in each frame is calculated as
\begin{equation}
\begin{aligned}
\label{eq::eq1}
A_{ij}^{t} = \exp\Big(-\dfrac{\left\|x_{i}^{t}-x_{j}^{t}\right\|_2}{\sigma}\Big),
\end{aligned}
\end{equation}
where $ x_{i}^{t}, x_{j}^{t} $ denote the feature descriptors of any two node pairs in current frame $t$ and $ \sigma $ is a scaling parameter. Similarly, we can also obtain $A^{t-1}$ for the previous frame.
Then, for the edges connected between different consecutive frames $B \in \mathbb{R}^{n_{t-1}\times n_t}$, we also use EPPM~\cite{Bao2014Fast} to extract the optical flow information  at the pixel level, as suggested in work~\cite{Yeo2017Superpixel}.
Finally, we obtain the spatial-temporal graph $A\in \mathbb{R}^{n\times n}, n=n_{t-1}+n_t$ of the two consecutive frames as
\begin{equation}
\label{eq::AA}
A=\left(\begin{array}{ll}{A^{t-1}} & {B} \\ {B^{T}} & {A^{t}}\end{array}\right)
\end{equation}
The constructed graph $G(V,E)$ is a symmetric graph.



\subsection{Graph  convolutional network}

Recently, graph convolutional network (GCN) has been commonly employed in many computer vision tasks, such as image co-saliency estimation~\cite{M-GLCN2019}, social influence prediction~\cite{Qiu2018DeepInf} and multi-label classification ~\cite{Multi-Label2019}, etc. In this paper, we adapt it for our tracking problem. Specifically, we formulate the problem of superpixel labeling as node labeling on the constructed superpixel graph $G(V,E)$ by employing a GCN model. In general, GCN consists of an input layer, several propagation layers and a final output layer. 
Given superpixel features $X\in \mathbb{R}^{n\times d}$ and graph adjacency matrix $A\in \mathbb{R}^{n\times n}$ Eq.(~\ref{eq::AA}), then we can use a two-layer GCN model proposed in work~\cite{KipfSemi} as
%
\begin{equation}
\begin{aligned}
\label{eq::GCN}
z=&\phi(X, A;W^{(0)},W^{(1)})  \\
=& \operatorname{softmax}\big(\hat{A} \operatorname{ReLU}\big(\hat{A} X W^{(0)}\big) W^{(1)} + \textbf{1}b\big)
\end{aligned}
\end{equation}
where $\hat{A}=D^{-\frac{1}{2}} A D^{-\frac{1}{2}}$ and $\textbf{1}=(1,1\cdots 1)$. Parameter $ W^{(0)}\in \mathbb{R}^{d \times h} $ represents the weight matrix of the first layer and $ W^{(1)}\in \mathbb{R}^{h \times 1}$ represents the weight matrix of the second output layer. Here, $h$ denotes the number of hidden layer units. $b$ is a bias term.
The final output $z\in \mathbb{R}^{n \times 1}$ denotes the final label prediction for all graph nodes, i.e., $z_i$ represents the confidence/weight that the $i$-th superpixel belongs to the target object.




To maintain more discriminative information, Park et al.,~\cite{ParkSymmetric} propose GALA (Graph convolutional Autoencoder using Laplacian smoothing and sharpening) for graph node feature representation.
Using GALA~\cite{ParkSymmetric}, we
%
%
propose a two-layer graph network for superpixel labeling as
\begin{equation}
\begin{aligned}
\label{eq::GCN}
z=&\phi(X, A;W^{(0)},W^{(1)}) \\
= & \operatorname{softmax}\big(\hat{A}_m \operatorname{ReLU}\big(\hat{A}_h X W^{(0)}\big) W^{(1)} + \textbf{1}b \big)
\end{aligned}
\end{equation}
Here,  $\hat{A}_m$ and $\hat{A}_h$ are defined respectively as~\cite{ParkSymmetric}
\begin{equation}
\begin{aligned}
\label{eq::mode1_our}
\hat{A}_m & = I - \lambda_{1}(I-D^{-\frac{1}{2}} A D^{-\frac{1}{2}}) \\
\hat{A}_h & =I + \lambda_{2}(I-D^{-\frac{1}{2}} A D^{-\frac{1}{2}})
\end{aligned}
\end{equation}
where $ D $ denotes a diagonal matrix with $ D_{ii} = \sum_{j} A_{ij} $ and $\lambda_1, \lambda_2$ are two parameters. 


In addition, to further simplify the nonlinear mapping between our GCN layers, as suggested in work~\cite{WuSimplifying}, we propose to transform the nonlinear the above nonlinear model into a simple linear model. 
Formally, by removing nonlinear function $\mathrm{ReLU}$, the above model becomes, 
%
\begin{equation}
\begin{aligned}
\label{eq::GCN1}
z& =\phi \big(X, A;W^{(0)},W^{(1)}\big)\\
& =\operatorname{softmax}\big(\hat{A}_m\hat{A}_h X W^{(0)}W^{(1)} + \textbf{1}b \big)
\end{aligned}
\end{equation}
It can be rewritten compactly as
\begin{equation}
\begin{aligned}
\label{eq::GCN2}
z=\phi(X, A;W)=\operatorname{softmax}\big(\hat{A}_m\hat{A}_h X W + \textbf{1}b \big)
\end{aligned}
\end{equation}
where $W = W^{(0)}W^{(1)}\in \mathbb{R}^{d\times 1}$.

In the following, we propose an effective way to train the proposed model~Eq.(\ref{eq::GCN2}). The optimization algorithm of loss function is summarized in Algorithm~\ref{alg:algorithm}.

\subsection{Loss function}

%
%
In this paper, we propose to learn the optimal network parameters $\{W, b\}$ by employing a semi-supervised learning manner.
Specifically, the already target tracked result in previous $t-1$ frame provides the labelled information to guide the training of the proposed model.
Formally, let $f=(f_1, f_2,\cdots f_{n_{t-1}})\in \mathbb{R}^{n_{t-1}\times 1}$ be the indicative vector of target tracked/segemented result in the previous frame, i.e.,
$f_i=1$ if the $i$-th superpixel belongs to target in previous $t-1$ frame, and  $f_i=0$ otherwise.
%
%
Then, we adopt the flexible manifold ranking model~\cite{wang2018flexible} to define our loss function as 
%
\begin{equation}
\begin{aligned}
\label{eq::Fusion Model}
& \mathcal{L}(y,W,b) =  \\
& \|\phi(X, A;W,b) -y\|^2_2
+ \alpha y^TL_Sy + \beta \|(y^{t-1}-f)\|_2^2
\end{aligned}
\end{equation}
where $ y=(y^{t-1},y^t) $ is a variable vector to be optimized and
$ L_S = D-A $ is the Laplacian matrix.
The first residual term denotes the label prediction loss.
The second term is the smoothing constraint which encourages neighboring nodes have similar labels.
The last term denotes the label fitting term.
The main benefit of introducing the variable $y$ is to make the model be more flexible, as suggested in work~\cite{wang2018flexible}.

\section{Optimization}

In this section, we propose an effective algorithm to optimize the proposed model~Eq.(\ref{eq::GCN2}).
For efficiency consideration,
we propose to optimize the loss~Eq.(\ref{eq::Fusion Model}) by ignoring the softmax function.
In this case, we can derive a simple update algorithm to optimize the loss function.
Model~Eq.(\ref{eq::GCN2}) contains three trainable parameters, i.e., network work parameters $\{W, b\}$ and variable $y$. They are optimized by alternatively optimize them until convergence, as summarized in Algorithm~\ref{alg:algorithm}.

\textbf{(1) $\{W, b\}$-problem: Fix $ y $, update $\{W, b\}$.}
The problem becomes
\begin{equation}
\begin{aligned}
\label{eq::solve Model_1}
& \min_{W,b} \|\phi(X, A;W,b) -y\|^2_2
\end{aligned}
\end{equation}
By ignoring the softmax function, it becomes

\begin{equation}
\begin{aligned}
\label{eq::solve Model_1}
\min_{W,b} \|(\hat{A}_m\hat{A}_h X)^{T}W +\textbf{1} b -y\|^2_2 
\end{aligned}
\end{equation}
%
It is known that, the optimal $ W,b $ can be acquired by setting the first order derivative of Eq.(\ref{eq::solve Model_1}) with respect to variables $ W,b $ to zeros respectively which is given as
\begin{equation}
\begin{aligned}
\label{eq::solve Model_11}
&W^{*}=\left((\hat{A}_m\hat{A}_h X)^{T}\right)^{-1}(X(\textbf{1} b-y)
\end{aligned}
\end{equation}
\begin{equation}
\begin{aligned}
\label{eq::solve Model_12}
&b^* =\frac{1}{n} \textbf{1}^T \big(y-(\hat{A}_m\hat{A}_h X)^{T}W\big)
\end{aligned}
\end{equation}

\textbf{(2) y-problem: Fix $\{W, b\}$, update $ y $.} The problem becomes
\begin{equation}
\begin{aligned}
\label{eq::solve Model_2}
& \min_{y} \|(\hat{A}_m\hat{A}_h X)^{T}W +\textbf{1} b -y\|^2_2 + \alpha y^TL_Sy + \beta \|y^{t-1}-f\|_2^2 \\
& s.t.\ \ y_i\geq 0
\end{aligned}
\end{equation}
which is simply rewritten as
\begin{equation}
\begin{aligned}
\label{eq::solve Model_21}
\min_{y} \|y - q\|_2^2 +\alpha y^TL_Sy \ \ \ \  s.t.\ \ y_i \geq 0
\end{aligned}
\end{equation}
where $q=\Gamma ((\hat{A}_m\hat{A}_h X)^{T}W +1 b +\beta f) $ and $ \Gamma = diag(u^{t-1}, u^{t}) $,
$ u^{t-1} $ = $[1, 1, ..., 1] \in \mathbb{R}^{n^{t-1} \times 1} $, $ u^{t} $ = $ [0, 0, ..., 0] \in \mathbb{R}^{n^{t} \times 1} $.
We can simply prove that Eq.(~\ref{eq::solve Model_21}) can also be written as
\begin{equation}
\begin{aligned}
\label{eq::solve Model_42}
\min_{y} \|y - \hat{q}\|_2^2 +\alpha y^TL_Sy 
\end{aligned}
\end{equation}
\noindent where $ {\hat{q}}_i = \max\{q_i,0\} $.
It is known that the Eq.(~\ref{eq::solve Model_42}) is a convex function, so its optimal solution must be non-negative. Thus, by setting the first derivative of Eq.(~\ref{eq::solve Model_42}) with respect to variable $ y $ to zero. Then the optimal $ y $ can be obtained as
\begin{equation}
\begin{aligned}
\label{eq::solve Model_43}
y^* = (I +\alpha L_S)^{-1}\hat{q}
\end{aligned}
\end{equation}

\begin{algorithm}[ht]
	\caption{Optimization algorithm of~(Eq.(\ref{eq::GCN2}))}
	\label{alg:algorithm}
	\textbf{Input}: The feature descriptor $X=(X^{t-1}, X^{t})$, $X \in \mathbb{R}^{d\times n}$, indicative vector $f \in \mathbb{R}^{n\times 1}$ and two kinds of graph $ \hat{A}_m$, $\hat{A}_h $. \\	
	\textbf{Parameter}: $ \lambda_{1}=0.01 $, $\lambda_{2}=0.07$, $\alpha=0.001$, $ \beta=50 $, $ minError=1e-4 $,
	$ maxLter=le-5 $. \\
	\textbf{Output}: Variable vector. $y\in\mathbb{R}^{n \times 1}$. \\
	\textbf{Initialize}: Indicator vector $f$ is initialized with the segmentation result of the previous frame, and	linear regression parameter $W,b$ are set to $\textbf{0}$ and $0$.	
	
	\begin{algorithmic}[1] 
		\WHILE{not converged}		
		\STATE Update $ W $ by solving the problem  Eq.(~\ref{eq::solve Model_11});
		\STATE Update $ b $ by solving the problem  Eq.(~\ref{eq::solve Model_12});	
		\STATE Update $ y $ by solving the problem  Eq.(~\ref{eq::solve Model_43});
		\ENDWHILE
		\STATE \textbf{until} converge
		\STATE \textbf{return} $ y $;
	\end{algorithmic}
\end{algorithm}

\section{Experiment}

To evaluate the effectiveness of the proposed GCN-based tracking approach. We
test it on several datasets and compare it with some other related approaches.
This section provides both quantitative and qualitative results of our proposed approach.
\subsection{Datasets}

We evaluate the proposed algorithm on five benchmark datasets
including DAVIS~\cite{Perazzi2016A}, GBS~\cite{6126494,Lim2014Generalized,Lim2013Joint}, ST2~\cite{Li2013Video}, NR~\cite{OGBDT2015} and VS~\cite{Fukuchi2009Saliency} datasets.
The detail of these datasets are introduced below.

\begin{itemize}
	\item DAVIS~\cite{Perazzi2016A} contains 50 different video sequences which covers the common challenges such as deformation, occlusion, and appearance changes in target segmentation and tracking tasks.
	\item GBS (Generalized Background Subtraction)~\cite{6126494,Lim2014Generalized,Lim2013Joint} dataset contains 15 video sequences which
	also includes deformation, occlusion, and low resolution.
	\item ST2 (SegTrack v2)~\cite{Li2013Video} dataset contains 24 targets which are  mainly for problems similar to GBS and some of them require tracking multiple ground truth objects.
	\item NR (Non-rigid object tracking)~\cite{OGBDT2015}  contains 10 challenging video sequences which are challenging mainly due to articulated and non-rigid deformable objects.
	\item VS (Video Saliency)~\cite{Fukuchi2009Saliency} dataset consists of 10 sequences with complex backgrounds or multiple objects which considers only the most significant objects and ignores those that are too small to segment.
\end{itemize}


We compare the proposed method with six state-of-the-art tracking-by-segmentation methods and four tracking-by-detection methods, including AMCT~\cite{Yeo2017Superpixel}, OGBDT~\cite{OGBDT2015}, HT~\cite{Godec2011Hough}, SPT~\cite{Shu2011Superpixel}, PT~\cite{Duffner2014PixelTrack},  MUSTer~\cite{Hong2015MUlti}, DSST~\cite{Danelljan2014Accurate} and MEEM~\cite{MEEM2014}. 

\subsection{Experimental Setup and Evaluation Metrics}

\begin{table*}
	\small
	\centering
	\caption{Average overlap ratio of segmentation masks for tracking-by-segmentation algorithms.}
	\label{table::table1}
	\setlength{\tabcolsep}{4.5mm}{
		\begin{tabular}{l | c| c | c | c | c | c  c }
			\hline
			\hline
			& Ours  & AMCT~\cite{Yeo2017Superpixel} & OGBDT~\cite{OGBDT2015} & HT~\cite{Godec2011Hough} &SPT~\cite{Shu2011Superpixel} & PT~\cite{Duffner2014PixelTrack}  \\
			\hline
			{DAVIS~\cite{Perazzi2016A}}  & {\bf 62.0}  & 59.2 & 44.9 & 33.1 & 27.1 & 26.1\\
			{GBS~\cite{6126494,Lim2014Generalized,Lim2013Joint}}  & {\bf 75.4}   & 74.8 & 59.7 & 40.4 & 45.9 & 35.3\\
			{ST2~\cite{Li2013Video}}  & {\bf 64.0}  & 58.8 & 47.6 & 43.0 & 26.3 & 21.2\\
			{NR~\cite{OGBDT2015}}  & {\bf 60.3}   & 58.6 & 53.3 & 41.1 & 29.7 & 28.3\\
			{VS~\cite{Fukuchi2009Saliency}}  & {\bf 87.6}  & 84.1 & 79.8 & 51.2 & 61.0 & 73.9\\
			\hline
			\hline
	\end{tabular}}
\end{table*}
\begin{table*}
	\small
	\centering
	\caption{Average overlap ratio of bounding boxes for tracking algorithms.}
	\label{table::table2}
	\vspace{0.1 ex}
	\setlength{\tabcolsep}{1mm}{
		\begin{tabular}{l | c  |c |c |c| c| c | c| c| c c}			
			\hline
			\hline
			& \multicolumn{6}{c|}{tracking-by-segmentation algorithms}& \multicolumn{4}{c}{tracking-by-detection algorithms} \\
			
			\cline{2-11}
			~ & Ours  & AMCT~\cite{Yeo2017Superpixel} & OGBDT~\cite{OGBDT2015} & HT~\cite{Godec2011Hough} & SPT~\cite{Shu2011Superpixel} & PT~\cite{Duffner2014PixelTrack} &  MUSTer~\cite{Hong2015MUlti} & DSST~\cite{Danelljan2014Accurate} & MEEM~\cite{MEEM2014}  \\
			\hline
			{DAVIS~\cite{Perazzi2016A}}  &{\bf 61.8}   & 60.9 & 50.0 & 35.8 & 43.2 & 41.6  & 25.9 & 58.4 & 52.7\\
			{GBS~\cite{6126494,Lim2014Generalized,Lim2013Joint}}  &{\bf 81.2}   & 80.0 & 61.2 & 43.0 & 55.2 & 44.7  & 59.4 & 62.9 & 52.6\\
			{ST2~\cite{Li2013Video}}  &{\bf 68.6}  & 64.8 & 50.2 & 44.9 & 53.5 & 32.2  & 58.8 & 62.0 & 59.5\\
			{NR~\cite{OGBDT2015}}  &{\bf 67.5}  & 66.9 & 60.8 & 40.9 & 35.7 & 16.1  & 36.2 & 35.4 & 33.1\\
			{VS~\cite{Fukuchi2009Saliency}}  &{\bf 90.6}  & 88.2 & 78.8 & 57.6 & 61.5 & 51.9  & 64.1 & 66.9 & 60.3\\
			\hline
			\hline
	\end{tabular}}
\end{table*}

\begin{figure*}
	\subfigure[]{
		\begin{minipage}[t]{0.33\textwidth}
			\centering
			\includegraphics[width=\textwidth]{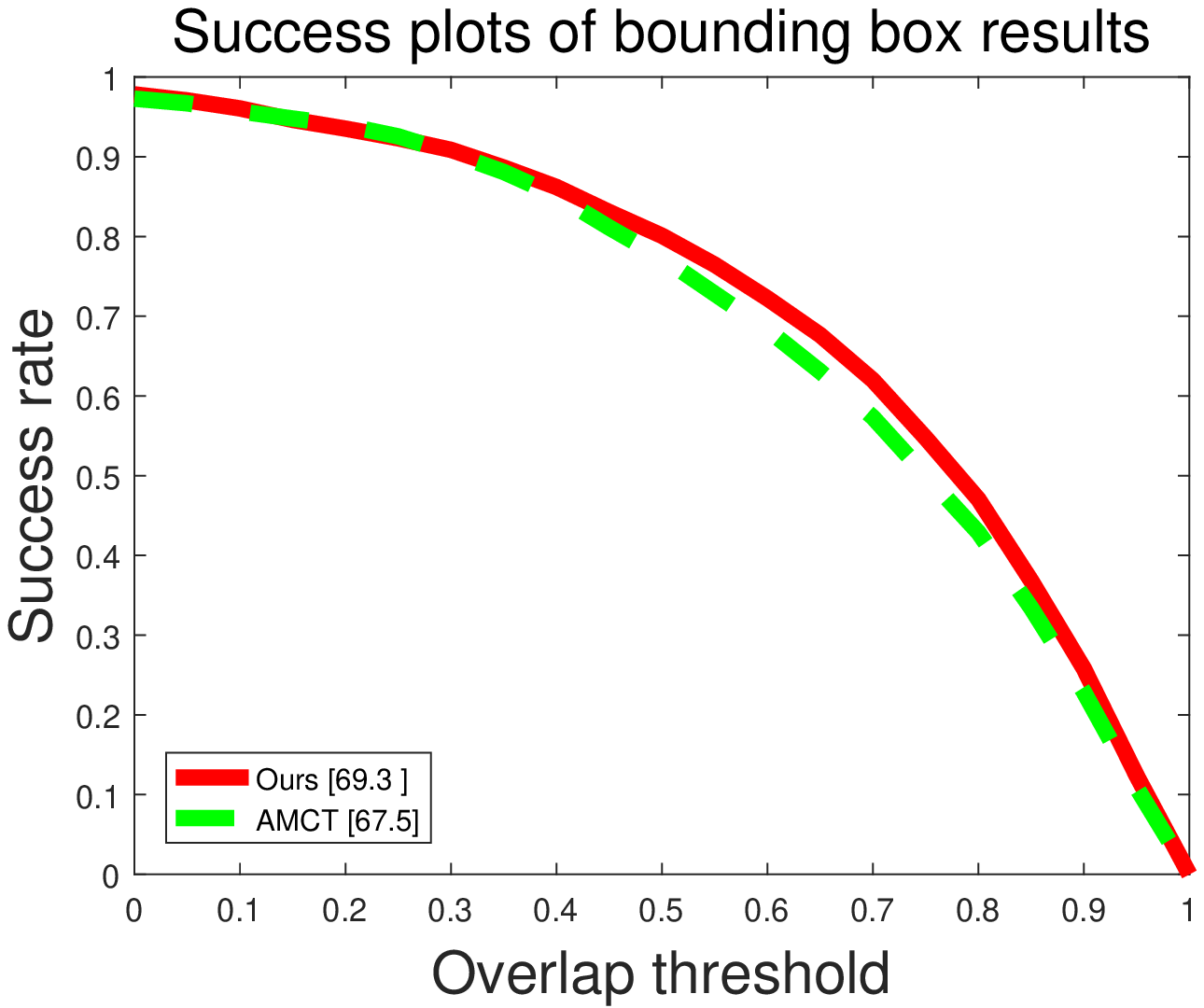}
		\end{minipage}
	}
	\subfigure[]{
		\begin{minipage}[t]{0.33\textwidth}
			\centering
			\includegraphics[width=\textwidth]{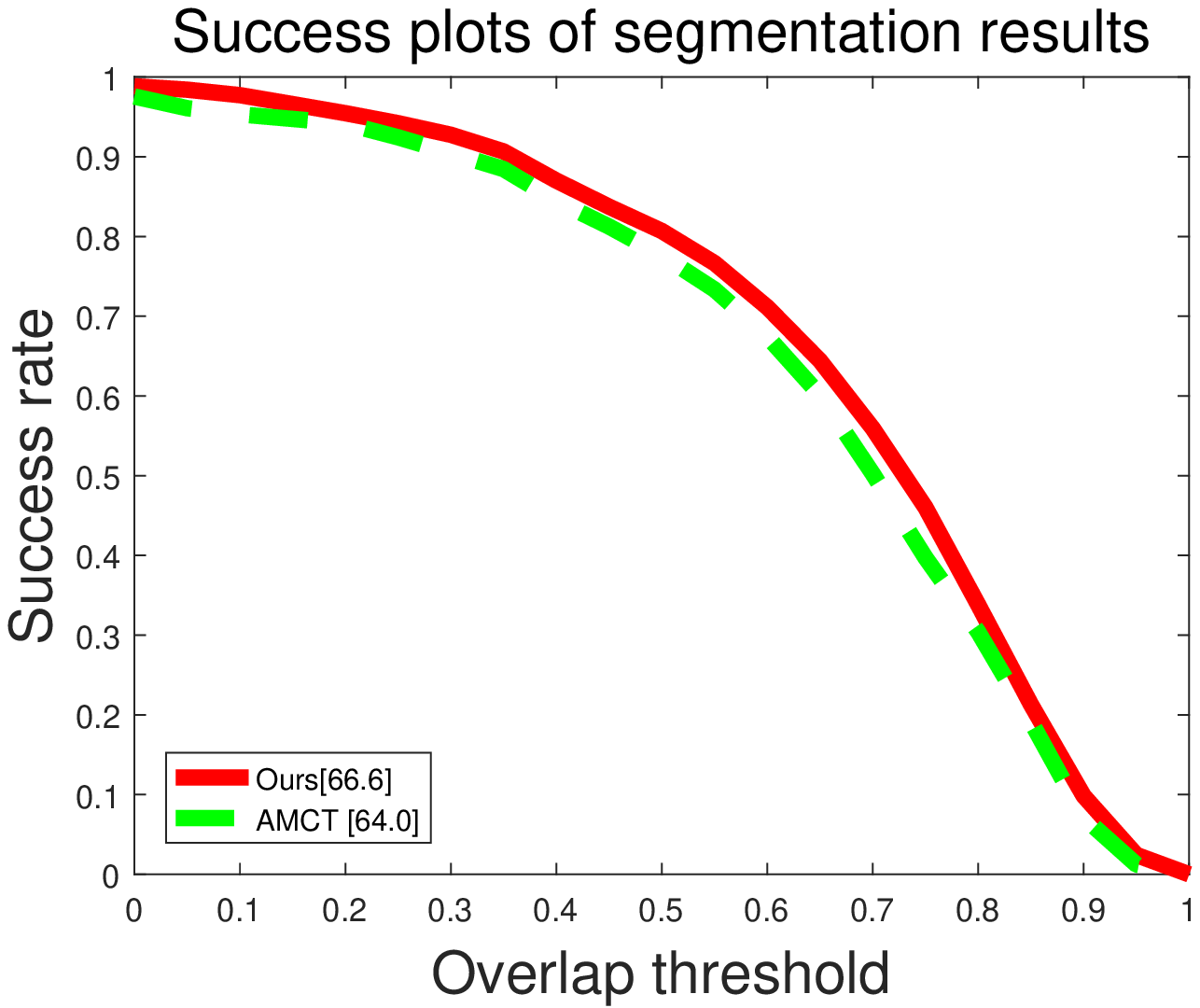}
		\end{minipage}
	}
	\subfigure[]{
		\begin{minipage}[t]{0.33\textwidth}
			\centering
			\includegraphics[width=\textwidth]{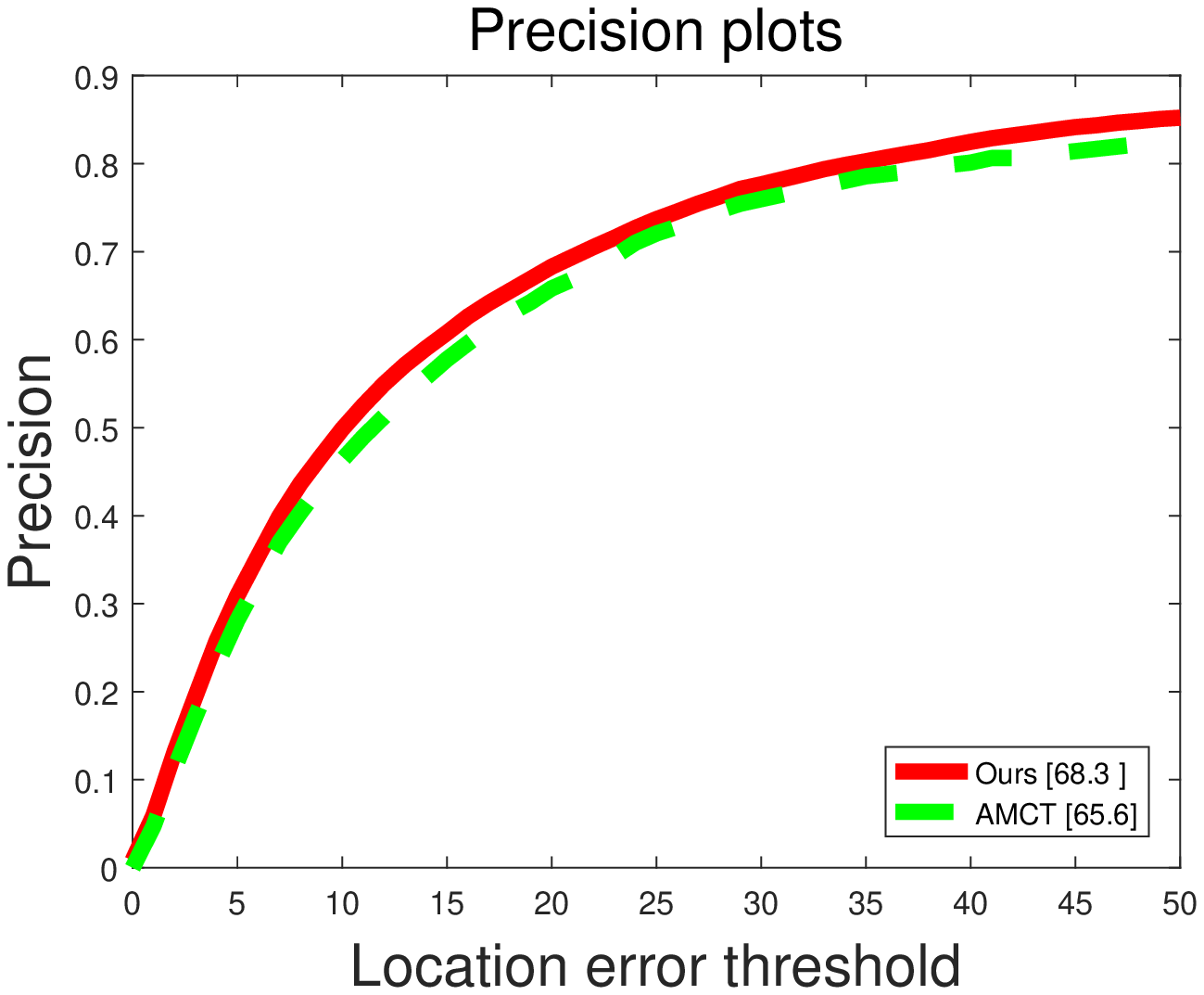}
			\label{img:precision}
		\end{minipage}
	}

	\caption{Success and precision plots on five datasets: (a) success plots in terms of  bounding box overlap ratio (b) success
		plots in terms of segmentation overlap ratio (c) precision plots.}
	\label{img::plots}
\end{figure*}
\begin{figure*}	
	\centering
	\includegraphics[scale=0.65]{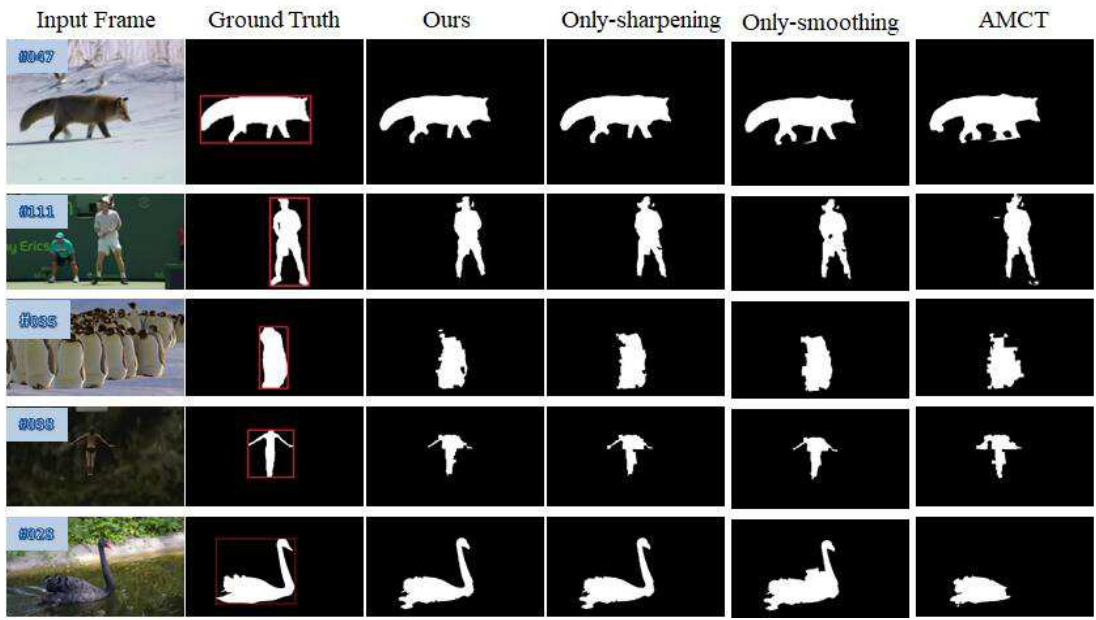}
	\caption{With challenges of non-rigid object deformation, occlusion,  appearance changes and other challenges. From top to bottoms, we show segmentation and tracking results of \emph{fox1} in VS, \emph{tennis} in GBS, \emph{penguin3} in ST2, \emph{Volleyball} in NR and \emph{blackswan} in Davis.}
	\label{img::demo}
\end{figure*}

\textbf{Parameter Settings.}
We partition each candidate region into about $600$ non-overlapping superpixels. Similar to experimental setting in work~\cite{Yeo2017Superpixel}, we extract 3-dimensional mean colors in LAB space features for each superpixel. The parameters in Eq.(\ref{eq::mode1_our}) are set empirically as $\left\{\lambda_{1} = 0.01, \lambda_{2} = 0.07 \right\}$. The parameters $\{\alpha,\beta\}$ in Eq.(\ref{eq::Fusion Model})) are set to $\left\{0.001, 50 \right\}$. In Eq.(\ref{eq::eq1}), we set the scaling parameter $ \sigma = 10 $. \\

\textbf{Evaluation Metrics.} Similar to many other tracking works, our tracking results are measured based on two evaluation metrics, i.e., distance precision rate and overlap success rate~\cite{Henriques2015High,OTB2015}.
For precision rate, it is defined to measure the difference between the estimated position and the center position of the ground-truth. The frame whose center distance is less than a certain threshold is regarded as the correctly tracked frame and participates in the precision calculation. The final precision is determined by the ratio of correct frames w.r.t total frames. We set the threshold to 20 pixels.
For overlap success rate, the average intersection over union (IoU) is used to measure overlap  between the tracking result and the ground-truth. We calculate the segmentation  overlap ratio (average IoU ratios of segmentation masks and ground-truth masks) and bounding box overlap ratio (average IoU ratios of ground-truth and estimated bounding boxes), respectively.
The results of comparison methods~\cite{Yeo2017Superpixel,OGBDT2015,Godec2011Hough,Shu2011Superpixel,Duffner2014PixelTrack,Hong2015MUlti,Danelljan2014Accurate,MEEM2014} have been reported in previous works~\cite{Yeo2017Superpixel,Tracking-by-Segmentation2018} and we use them in Table~\ref{table::table2}.


\subsection{Performance Evaluation}

We compare the proposed model against three traditional  tracking-by-detection methods, namely MUSTer~\cite{Hong2015MUlti}, DSST~\cite{Danelljan2014Accurate} and MEEM~\cite{MEEM2014}, and five segmentation-based tracking algorithms including AMCT~\cite{Yeo2017Superpixel}, OGBDT~\cite{OGBDT2015}, HT~\cite{Godec2011Hough}, SPT~\cite{Shu2011Superpixel}, PT~\cite{Duffner2014PixelTrack}.\\

Table~\ref{table::table1} reports the results of average IoU ratios of segmentation masks and ground truth masks on all datasets. We can note that the proposed method exceeds some other tracking-by-segmentation methods, especially on ST2 data-set~\cite{Li2013Video}. This clearly demonstrates the effectiveness of the proposed GCN tracking model by further incorporating structural information into superpixel feature representation and thus can obtain more accurate segmentation results.
The proposed method can also obtain better performance than other methods on some larger datasets, such as  DAVIS~\cite{Perazzi2016A}.
%
%
Table~\ref{table::table2} shows average IoU ratios of ground truth and estimated bounding boxes on different datasets. The first column in the table illustrates that our method not only outperforms existing tracking-by-segmentation methods, but also obviously performs better than some other tracking-by-detection methods. The comparsion results of these methods have been reported in ~\cite{Yeo2017Superpixel}, here, we use them. 
We also compare our method with AMCT~\cite{Yeo2017Superpixel} on the area under curve (AUC) which is most related with our approach.
Figure~\ref{img::plots} (a) and (b) show the success rate in terms of bounding box and segmentation mask, respectively.
Figure~\ref{img::plots} (c) shows the precision plots in terms of bounding box.
Figure~\ref{img::demo} 
It can be intuitively obtained from Figure~\ref{img::plots}, bounding box overlap ratio performance exceeds AMCT~\cite{Yeo2017Superpixel} by 1.8\% and 2.6\% higher in the segmentation overlap ratio. Compared with  AMCT~\cite{{Yeo2017Superpixel}},  the proposed method can gain the precision of the bounding boxes gains of 2.7\%. Figure~\ref{img::demo} shows some visualization results.

\begin{table}[h]
	\small
	\centering
	\caption{Comparison of the proposed algorithm with only-smoothing, and none of both on overall datasets.}
	\label{table::table3}
	\vspace{1 ex}
	\setlength{\tabcolsep}{1.0mm}{
		\begin{tabular}{l|c|c|c}
			\hline
			\hline
			& Ours & only-smoothing  & none \\
			\hline
			{Success-Seg} & {\bf 66.6} & 65.6 & 64.6  \\
			{Success-Box} & {\bf 69.3} & 68.2  & 67.1  \\
			{Precision} & {\bf 68.3} & 68.1  & 66.7 \\			
			\hline
			\hline
	\end{tabular}}
\end{table}

\subsection{Ablation study}

In this section, we will focus on the detailed analysis of the proposed GCN feature learning module. The general graph convolution is the same as Laplacian smoothing.
As shown in table~\ref{table::table3}, we explore the effects of Laplacian smoothing and our moudle on the experiment respectively. Among them, only-smoothing means that the features are only learned by employing one layer graph Laplacian smoothing convolution. 'none' means that we do not perform any graph convolutional operations on the original input features. 'Success-Seg', 'Success-Box' and 'Precision' correspond to the three evaluation results in Figure~\ref{img::plots}.
Here, we can note that, the proposed graph mixed convolutional network model performs more effectively than purely using smoothing based graph convolution. That is because the proposed GCN model fuses the two convolutional operations and thus fully considers the discriminability and similarity of the features. The best results are shown in bold in the table~\ref{table::table3}, which obviously validates the effectiveness of the proposed model.
%

\section{Conclusion}

This paper proposes a simple and efficient graph convolutional representation model for object segmentation-based tracking tasks. The proposed model provides a general end-to-end framework which integrates label linear prediction and structure-aware feature information of each super-pixel together to obtain object segmentation and further improve the performance of tracking. An effective algorithm has been developed to optimize the proposed model. Extensive experiments show that our method obtains better performance.

{\bibliographystyle{elsarticle-num}}
\bibliography{Reference}

%
%
%

\end{document}